\documentclass[10pt,twocolumn,letterpaper]{article}

\usepackage{iccv}
\usepackage{times}
\usepackage{epsfig}
\usepackage{graphicx}
\usepackage{amsmath}
\usepackage{amssymb}

\usepackage[pagebackref=true,breaklinks=true,letterpaper=true,colorlinks,bookmarks=false]{hyperref}

\iccvfinalcopy 

\def\iccvPaperID{4321} 

\ificcvfinal\fi

\begin{document}

\title{GANs N' Roses: Stable, Controllable, Diverse Image to Image Translation (works for videos too!) }

\author{Min Jin Chong and David Forsyth\\
University of Illinois at Urbana-Champaign\\
{\tt\small \{mchong6, daf\}@illinois.edu}
}

\maketitle

\begin{abstract}

We show how to learn a map that takes a content code, derived from a face image, and a randomly chosen style code to an anime image. We derive an adversarial loss from our simple and effective definitions of style and content. This adversarial loss guarantees the map is diverse -- a very wide range of anime can be produced from a single content code. Under plausible assumptions, the map is not just diverse, but also correctly represents the probability of an anime, conditioned on an input face. In contrast, current multimodal generation procedures cannot capture the complex styles that appear in anime.  Extensive quantitative experiments support the idea the map is correct. Extensive qualitative results show that the method can generate a much more diverse range of styles than SOTA comparisons. Finally, we show that our formalization of content and style allows us to perform video to video translation without ever training on videos. Code can be found here \url{https://github.com/mchong6/GANsNRoses}.

\end{abstract}

\begin{figure*}[ht]
    \centering
    \includegraphics[width=1\linewidth]{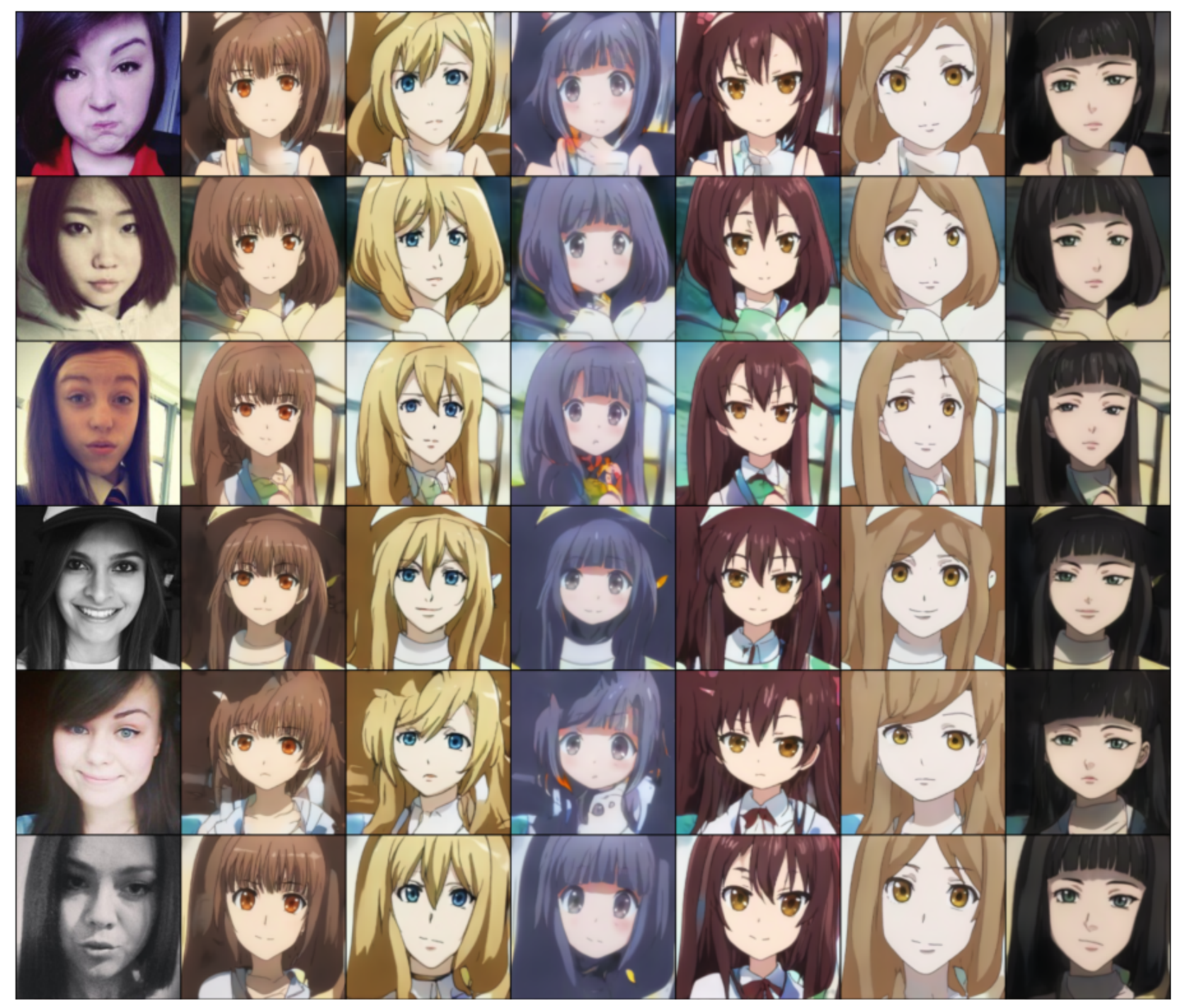}
    \caption{
      Our loss forces GNR to generate {\bf diverse outputs} from randomly sampled styles, by requiring that anime generated from augmented versions of a single image generate a batch that is indistinguishable from a sample of the same size of $P(Y)$.  As this figure shows, GNR can produce highly diverse outputs.  Each row shows anime generated from the content code of the first image, using sampled style codes; the style codes
      are the same across each column.  The six different style codes shown here produce strongly different anime figures.  Note that eye shape, size, style and color change with change of style code (row);
      Similarly, chin shape, nose rendering, hair color, haircut, and so on vary.  For a given style code (column), there is strong consistency in these details.  The tilt of the head, overall shape of the face,
      and positioning of facial features, and fringe are all controlled by the content (first image in each row), as is the collar style (open neck is preserved (row 1), as is hoodie (r2); formal collar (r3); and scoop neck (r4)). Note the spatial coherence of the anime images -- we do not have (say) images with eyes different size, etc. (compare figure \protect \ref{fig:comparisons}).}
    \label{fig:results}
\end{figure*}

\section{Introduction}
Imagine building a mapping that takes face images and produces them to anime drawings of faces. Some parts -- the
content -- of the image may be preserved, but others -- the style -- must change, because the same face could be
represented in many different ways in anime. This means we have a one-to-many mapping, which can be represented as a
function that takes a content code (recovered from the face image) and a style code (which is a latent variable), and
produces an anime face. But there are important constraints that must be observed. We want {\bf control}: the content of
the anime face can be changed by changing the input face (for example, if the person turns their head, so should the
anime). We want {\bf consistency}: different real faces rendered into anime using the same set of latent variables
should clearly match in style (for example, if the person turns their head, the anime doesn't change style unless the
latent variables change). Finally, we want {\bf coverage}: every anime image should be obtainable using some combination
of content and style, so that we can exploit the full range of possible anime images.     

Our method -- GANs N' Roses or GNR -- is a multimodal I2I framework that uses a straightforward formalization of the
maps using style and content (section~\ref{framework}). Achieving our goals requires carefully structured losses
(section~\ref{objectives}). The most important step is to be exact about what is intended by content and what is
intended by style. We adopt a specific definition: content is what changes when face images are subject to a family of
data augmentation transformations, and style is what does not change. This definition is very powerful. Our data
augmentations involve scaling, rotating, cropping, \etc. Thus, the definition means that content is (in essence) where
parts of the face are in the image and style is how the parts of the face are rendered. 

This definition allows us to learn a mapping from face images to content codes. We then pass the content codes to a
decoder, which must produce anime from them and a style latent variable. It is also very important that the anime
produced from a given content code ``know'' what that code is, and we use a decoder to recover the code from the
anime. But one face should yield many anime faces, and one anime face should have the same content as many faces. This
means we cannot require 1-1 loop closure of images (in contrast to CycleGAN), but must close the loop on content codes
instead.    This creates a difficulty, as the method might try to ignore the style code to obtain better cycle
consistency on content.  We show how to ensure the method produces the correct distribution of anime for a given
content, resulting in a method that can produce very diverse anime. This diversity is important in applications; for example, a user might wish to
exploit the control that our style code offers (Fig.~\ref{fig:sefa}) to get an avatar with just the right eye shape or color.
Our contributions are: 
\begin{itemize}
    \item Our definition of content and style is easily operationalized; we show how to use it to ensure that the anime
      produced from a single content code are properly diverse.
      \item The resulting method is very effective at producing controllable, diverse synthesis, by both quantitative
        and qualitative measures.
    \item Our definition of content means our method can synthesize anime videos from face videos without ever being
      trained on multiple frames.
      \item While we describe our method in the context of face to anime translation (because anime have a particular
        rich variation in style that is difficult to capture, section~\ref{related}), the method applies to any
        translation problem.
\end{itemize}
All figures are best viewed in color at high resolution.


\section{Related Works}\label{related}
\paragraph{Image to Image Translation}
Image to image translation (I2I) involves learning a mapping between two different image domains. In general, we want the translated image to maintain certain image semantics from the original domain while obtaining visual similarities to the new domain. Early works on I2I involves learning a deterministic mapping between paired data~\cite{wang2018high,isola2017image_patchgan}. This was later extended to a multimodal mapping in BicycleGAN~\cite{zhu2017toward}. However, due to the limited availability of paired data, this approach is cannot scale up to bigger unpaired datasets. The pioneering work of CycleGAN~\cite{zhu2017unpaired} solves this problem by employing the use of cycle consistency to learn image to image translation for unpaired data. Following works~\cite{kim2019u,yi2017dualgan,kim2017learning} have used similar approach. A significant limitation of these works is the lack of diversity of the output images due to their unimodal mapping. This is inherently limiting as image-to-image translation is generally a multimodal problem. Recent works on multimodal translation have then expanded on this. MUNIT~\cite{huang2018multimodal} and DRIT~\cite{lee2018diverse} decomposes an image into a domain invariant content code and a domain-specific style code and employ random style sampling to produce diverse outputs. StarGANv2~\cite{choi2019stargan2} employs a single generator to produce diverse images for multiple domains. Mode Seeking GAN~\cite{mao2019mode} builds on top of DRIT and encourages output diversity by penalizing output images that are similar to each other when their input style codes are different. 

In general, current multi-modal frameworks lack a proper definition of style and content; it is unclear what exactly they each constitute. Also, visual inspection of their outputs reveals mode collapse. For a given image, the multiple outputs look very similar, often with just color and slight stylistic changes. One recent work in CouncilGAN~\cite{nizan2020council} enables diverse outputs by collaborating between multiple GANs. However in the difficult setting of selfie2anime, CouncilGAN cannot capture the complex artistic style of animes, collapsing to few modes and not expressing the stylistic diversity we expect. Very recently, AniGAN~\cite{li2021anigan} proposes new normalizations to allow selfie2anime by transferring color and textual styles while maintaining global structure. AniGAN generates multi-modal outputs based on reference images. Like previous methods, AniGAN does not have explicit style diversity and lack output diversities.

In contrast to previous work, GNR focuses on achieving controllable style diversity by using our simple but effective definition and content. Our translation is also robust, allowing it to be applied to video to video translation at no additional cost.

\section{GANs N' Roses}
\subsection{Framework}\label{framework}
\begin{figure}[t!]
    \centering
    \includegraphics[width=1\linewidth]{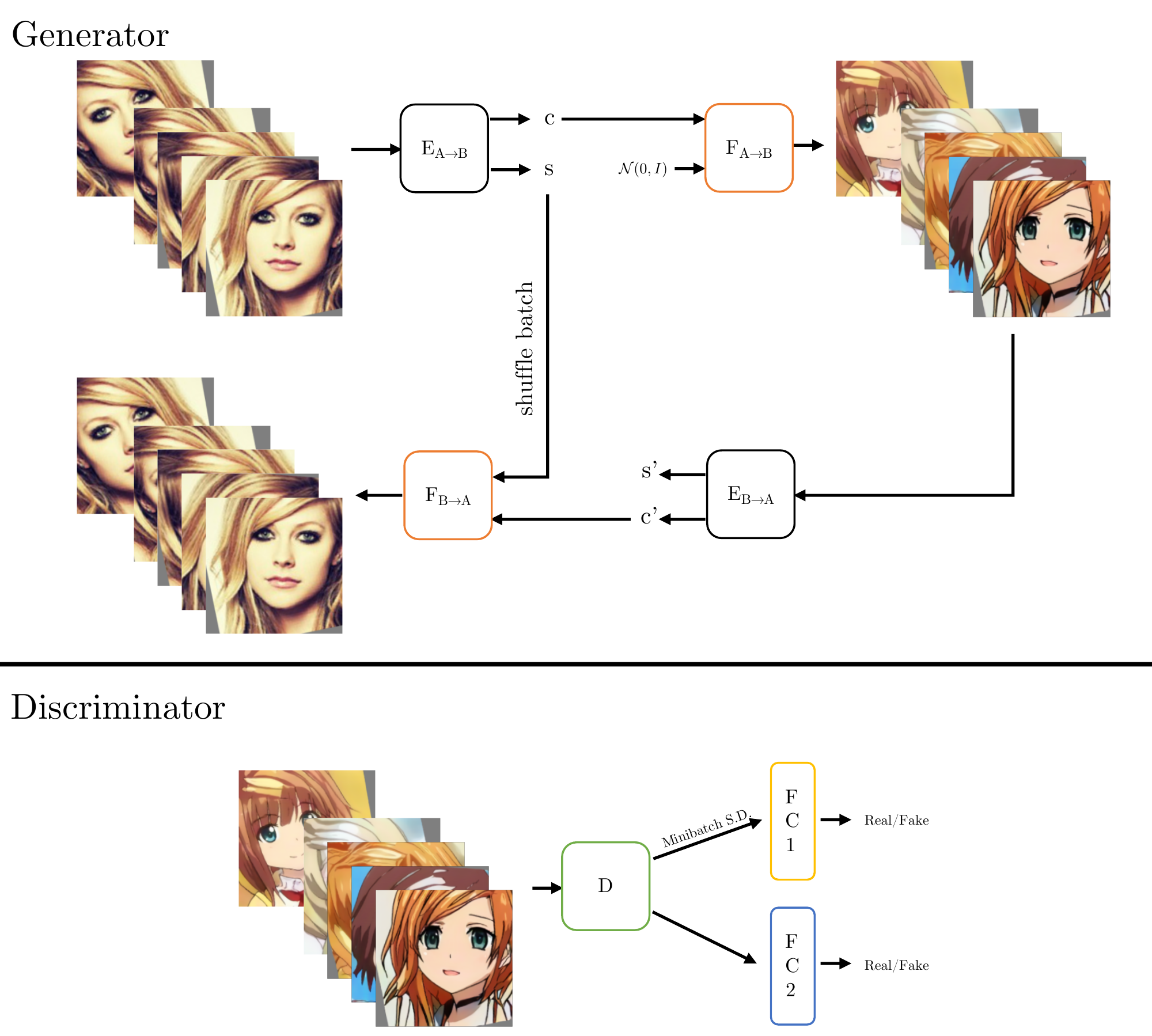}
    \caption{
    {\bf GANs N Roses Framework:}
    Unlike previous work which samples different images in a batch, GNR use the same image with different augmentations to form a batch. This allows us to constrain the spatial invariance in style codes \eg all style codes are the same across the batch. Our Diversity Discriminator looks at batch-wise statistics by explicitly computing the minibatch standard deviation across the batch. This ensures diversity within the batch.}
    \label{fig:framework}
\end{figure}

\begin{figure*}[t!]
    \centering
    \includegraphics[width=1\linewidth]{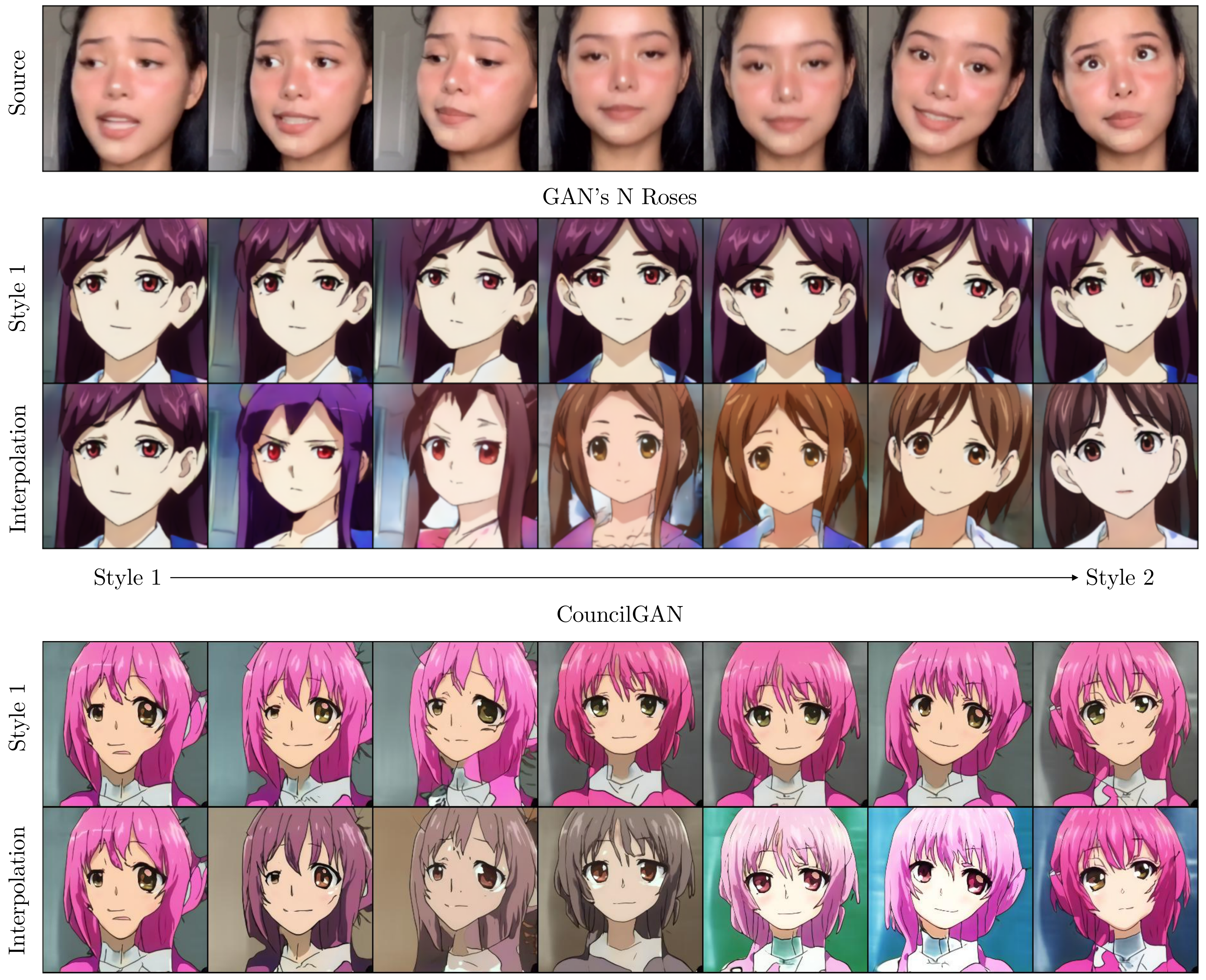}
    \caption{
    {\bf Video Temporal Coherence and Interpolation:}
    We show that our style code is temporally coherent in videos. We can also smoothly interpolate between styles as the face moves across time. In comparison, CouncilGAN introduces artifacts temporally with varying hair/background shades, uneven eyes and jaw shapes. The style outputs are also not diverse, and interpolation offers little changes except for color. Note that neither GNR nor CouncilGAN are trained on video dataset, but GNR perform significantly better due to our sensible definition of style and content. The full video is provided in the supplementary.}
    \label{fig:interpolation}
\end{figure*}

Given two domains $\mathcal{X}$ and $\mathcal{Y}$, for a  given $x \in \mathcal{X}$, our goal is to generate a diverse
set of $\hat{Y}$ in the $\mathcal{Y}$ domain that contains similar semantic contents with $x$. We expound 
translation from domain $\mathcal{X}$ to $\mathcal{Y}$ in detail (but exclude the other direction, which is mirrored, for brevity).   As Figure~\ref{fig:framework} shows, GANs N' Roses is made up of
an Encoder and a Decoder for each direction of $\mathcal{X} \rightarrow \mathcal{Y}$ and $\mathcal{Y} \rightarrow \mathcal{X}$. The encoder $E$ disentangles an image $x$ into a content code $c(x)$ and a style code $s(x)$. The
decoder $F$ takes in a content code and a style code and produces the appropriate image from $\mathcal{Y}$.  Together, the encoder and decoder form a
Generator.  At run time, we will use this generator by passing an image to the encoder, keeping the resulting content
code $c(x)$, obtaining some {\em other} relevant style code $s_z$, then passing this pair of codes to the decoder.
We want the content of the resulting anime to be controlled by the content code, and the style by the style code.  

But what is content, and what style?   The key idea of GANs N' Roses to define content as \emph{where
  things are} and style as \emph{what they look like}.  This can be
made crisp using the idea of data augmentations.  Choose a collection
of relevant data augmentations: style is all that is invariant under any of these, content is all that is not.  Note
that this definition is conditioned on the augmentations -- different sets of data augmentations will result in
different definitions of style.

\subsection{Ensuring Style Diversity}\label{diversity}

The framework of Fig.~\ref{fig:framework} is well established; the difficulty is ensuring that one actually gets very
different anime.  As section~\ref{related} sketches, there are three kinds of strategy in the current  literature.
First, one could simply generate from randomly chosen style codes $s_z$; but there is no guarantee of diversity -- the decoder might cheat, by simply ignoring
the style code, and produce only one style per face.  Second, one could  require that the decoder have the 
property that $s_z$ can be recovered from the decoder; but there is no guarantee of diversity -- the decoder might
cheat, by (say) hiding $s_z$ in a few pixel values, or by only changing the overall color of the image to signal $s_z$.
Third, one could write an explicit penalty that forces decodes with different style codes to be different; but there is
no guarantee the diversity is right -- the decoder might cheat by just changing the overall color of the decodes to be
different.  None of these strategies is satisfactory.

Our definition of style and content offers a cure. We must learn a map $F(c, s; \theta)$ that takes content codes $c$
and style codes $s$ to generate anime faces. Write $x_i \in \mathcal{X}$ for a single image, chosen uniformly at
random from the data, $T(\cdot)$ for a function that applies a randomly chosen augmentation to that image, $P(C)$ for
the distribution of content codes, $P(Y)$ for the true distribution of real anime (etc), and $\hat{Y}$ for generated anime.
We must have that $c(x_i) \sim P(C)$.  Because we define style to be what does not change under
augmentation, and content to be what does, reasonably selected augmentations should mean that $c(T(x_i)) \sim P(C)$ -- ie applying a random augmentation to an image
results in a content code that is a sample from the prior on content codes.  This assumption is reasonable -- if it was
strongly violated, then image augmentation for training classifiers would not work. 

At run time, the decoder is presented with content codes obtained by encoding true face images, and style codes that are
samples from some known distribution. Now consider a generated batch obtained by: (a) choose a single real face image;
(b) apply a random selection of augmentations to that image; (c) compute the content codes for those images, to obtain
$c_i$; (d) sample style codes  $z_i$ from $P(S)$, and generate $\hat{Y}_{i}=F(c_i, z_i; \theta)$. Because $C$ and $S$
are disentangled, they should be independent.  In turn, $(c_i, z_i) \sim P(C) P(S)$ and so
$\theta$ is correct when generated batches $\hat{Y}_{i}$ are indistinguishable from fair samples of $P(Y)$ that are the
same size.

Now assume that the generator has been trained such that, under the conditions above, generated batches $\hat{Y}_{i}$ are indistinguishable from fair samples of $P(Y)$ that are the
same size.  Because $P(Y|C=c_i)\propto P(Y, C=c_i)$, our model of $P(\hat{Y}|C=C_i)$ must be correct.  
More formally, write $T_n(\cdot)$ for a function that randomly chooses $n$ augmentations from the relevant family, and
applies them to the image.  We obtain a batch of content codes as $\left\{x_1, x_{\cdots}, x_n\right\} = T_n(x)$, then
form $(c_i, s_i) = E(x_i)$.  We then draw $z_i$ from $P(S)$, and form a batch of style content pairs
${\cal B}=\left\{(c_1,z_1), \ldots, (c_n, z_n)\right\}$.  We then use an adversary to ensure that the batch of $\hat{Y}_i=F(c_i,
z_i;\theta)$ is indistinguishable from a batch of $Y$.

\subsection{Losses}\label{objectives}

\paragraph{Style Consistency Loss}
The style code of an image must be invariant under $T(\cdot)$, so in a batch ${\cal B}$ as above we expect all $s_i$ to
be the same.  We use:
\begin{equation}
    \mathcal{L_{\text{scon}}} = \mathrm{Var}(s)
\end{equation}

\paragraph{Cycle Consistency Loss}

We want to ensure that the anime preserve the content code.  Generate an anime $\hat{y_i}$ from an image $x_i$ using
$(c(x_i), z_i)$, then map that anime back to an image using $(c(\hat{y_i}), s_i(x_i))$ (where the encoder for the content
code is the $\mathcal{Y}\rightarrow\mathcal{X}$ encoder).  We should get the original image back, because we are using
the original image's style, and a content code that should be preserved.  We find it helpful to shuffle the style codes
in a batch (all images in the batch ${\cal B}$ should have the same style code, above), because this encourages style consistency.  Writing
\begin{align}
\begin{split}
  &c'_i, s_i' = E_{\mathcal{Y} \rightarrow \mathcal{X}}(\hat{y_i})\\
    &\hat{x_i} = F_{\mathcal{Y} \rightarrow \mathcal{X}}(c'_i, s_j)
\end{split}
\end{align}
where $i \neq j$ and $\hat{x_i}$ is the reconstructed image in the original domain, the cycle consistency loss is:
\begin{equation}
    \mathcal{L_{\text{cyc}}} = \mathbb{E}_{x}\Big[\lVert \hat{x_i} - x_i\rVert_2\Big]
\end{equation}
In addition to L2 loss, we also apply LPIPs perceptual loss~\cite{zhang2018perceptual}.

\begin{figure*}[ht]
    \centering
    \includegraphics[width=1\linewidth]{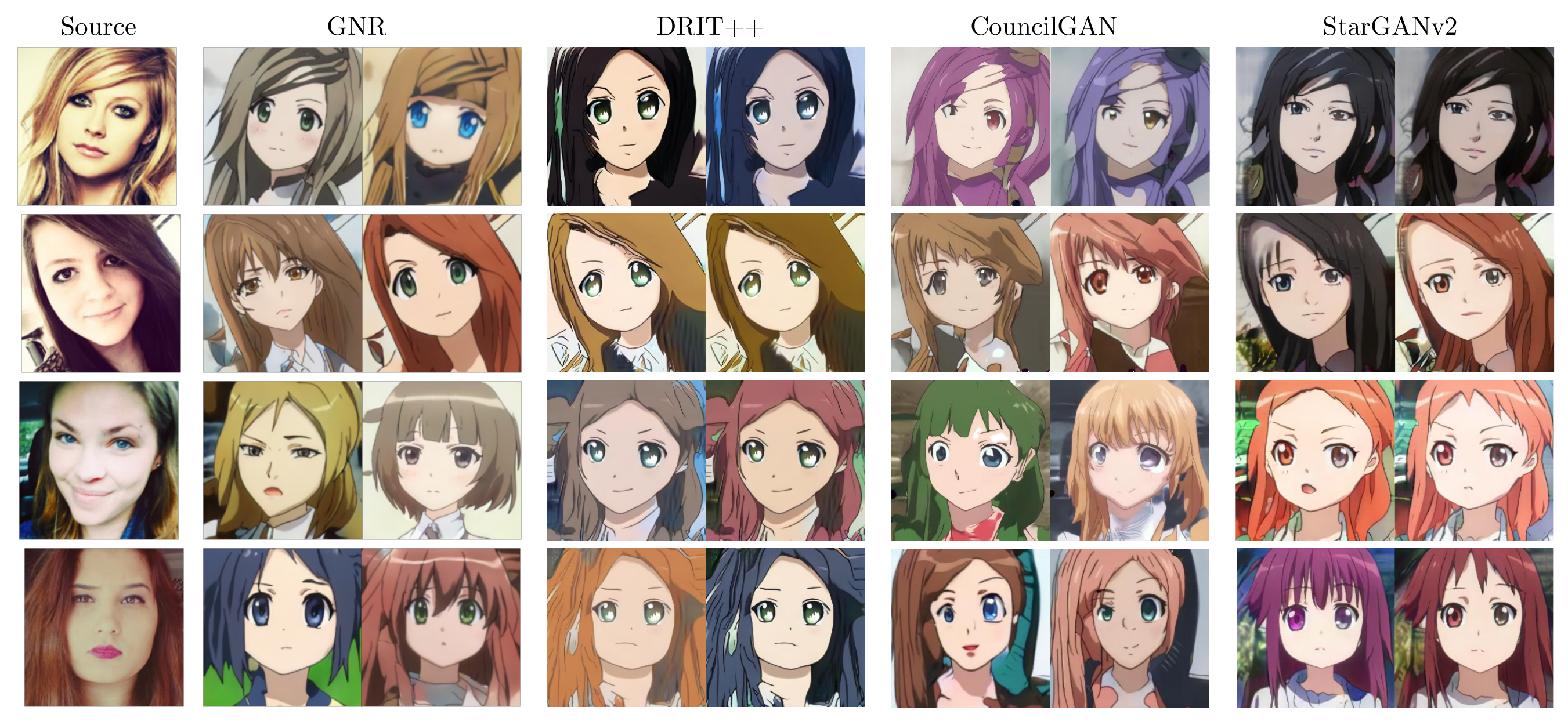}
    \caption{
    {\bf Comparisons:}
    We compare multi-modal results with SOTA translation frameworks. The two columns for each method correspond to different styles for the same source image, and thus, should look different -- this is only true for GNR. GNR produces significantly more diverse images compared to others, with varying eye sizes, hair styles, art style, face sizes, \etc. Notably, for DRIT++, the eyes very similar across all images, even for different source images; for CouncilGAN while there is slight variations in texture, the overall structure is almost identical across different styles; for StarGANv2, similar to CouncilGAN.  Note CouncilGAN's difficulty ensuring left and right eye are the same size; StarGANv2's floating fringe (row 2); CouncilGAN and DRIT++ get strong diversity in hair color, but StarGANv2 does not.}
    \label{fig:comparisons}
\end{figure*}

\begin{figure*}[t]
    \centering
    \includegraphics[width=1\linewidth]{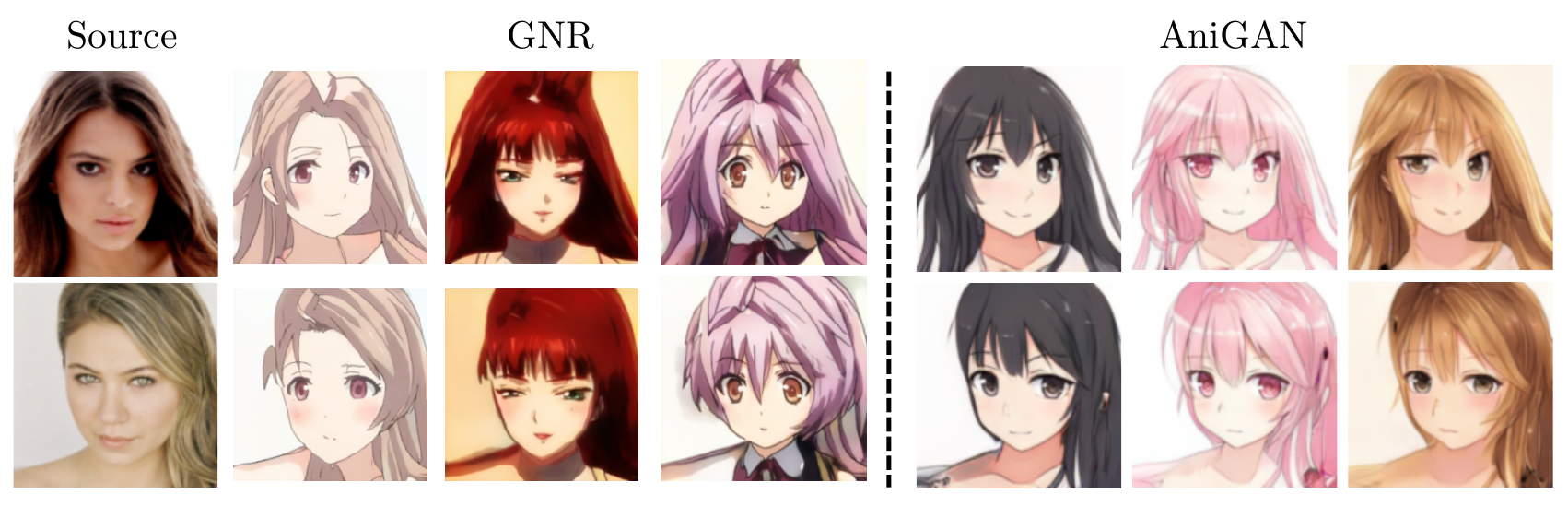}
    \caption{
    {\bf Comparisons:}
    We compare GNR with SoTA AniGAN. GNR produces much greater style diversity (eye shapes, hairstyles, line styles, \etc) while AniGAN produces very similar images that only changes hair color across style. Note that AniGAN was trained on a larger and less diverse dataset compared to GNR. AniGAN figures are taken from Fig.1 of their paper since their code is not yet released.}
    \label{fig:anigan}
\end{figure*}

\paragraph{Diversity Discriminator and Adversarial Loss}

\begin{figure*}[t!]
    \centering
    \includegraphics[width=1\linewidth]{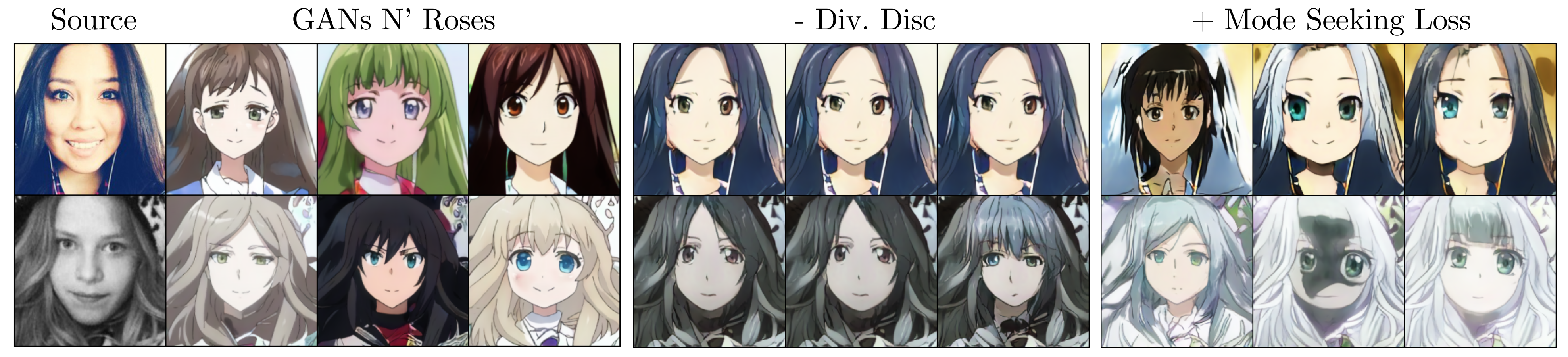}
    \caption{
        {\bf Ablation:} We compare GNR with 1) GNR without Diversity Discriminator 2) GNR without Diversity Discriminator and with Mode Seeking Loss~\cite{mao2019mode}. Style codes are randomly sampled for each column and not cherry picked. Diversity Discriminator clearly pushes GNR to produce more diverse and realistic outputs. Mode seeking loss introduces less diversity and encourages artifacts.}
    \label{fig:ablation}
\end{figure*}

Our criterion (a generated batch $\hat{Y}_{ij}$ be indistinguishable from a fair sample of $P(Y)$ of the same size)
applies to batches, not just individual samples. This means that we expect that across-sample properties computed within
batches should match (rather than just individual sample properties). At run time, conventional classifiers almost always are presented with individual images, rather than batches.  A
discriminator in an adversarial method is different, because it is trained with and applied to fixed batch sizes.  
Our  Diversity Discriminator exploits the minibatch standard  deviation trick used in PGGAN~\cite{karras2017progressive}
to exploit this difference. In the penultimate layer
of the discriminator, we compute the standard deviation for each feature and pass that into another FC layer. Our
Diversity Discriminator outputs the real/fake logits and the standard deviation logits. This means the discriminator can
identify reliable differences in within batch variation. We apply the non-saturating adversarial loss
$\mathcal{L_{\text{adv}}}$ from \cite{goodfellow2014generative} with R1 regularization~\cite{mescheder2018training} for 
both discriminator  branches.   Ablations show this diversity discriminator is important (Fig.~\ref{fig:ablation}),
likely because it increases discriminator efficiency: it is easier for a discriminator to spot subtle differences
between two sources if it sees whole batches.  We use

\paragraph{Total Loss}
Our total loss is 
\begin{equation}
        \mathcal{L} = \lambda_{\text{adv}}\mathcal{L}_{\text{adv}} + \lambda_{\text{scon}}\mathcal{L}_{\text{scon}} + \lambda_{\text{cyc}}\mathcal{L}_{\text{cyc}}
\end{equation}

\section{Experiments}
In all experiments, we use batch size of 7, $\lambda_\text{scon}=10, \lambda_\text{cyc}=20, \lambda_\text{adv}=1$. Our architecture is based on StyleGAN2~\cite{karras2020analyzing} with style code having a dimension of $8$. We use Adam optimizer~\cite{kingma2014adam} with a learning rate of $0.002$ for $300$k batch iterations for all networks. The random augmentations we use on the input images consist of random horizontal flip, rotation between $(-20, 20)$, scaling $(0.9, 1.1)$, translation $(0.1, 0.1)$, shearing $(0.15)$. The images are then upscaled to $286 \times 286$ and randomly cropped for resolution of $256 \times 256$. For datasets, we primarily focus on the selfie2anime dataset~\cite{kim2019u} with additional experiments of AFHQ~\cite{choi2019stargan2}.  We compare GANs N' Roses with several open source state-of-the-art multimodal I2I frameworks which we if available, use their pretrained models, otherwise, train using their default hyperparameters.

\subsection{Qualitative Comparisons}

Generally GNR produces strongly diverse images when given the same source image and different random style codes. The
style codes drive appearance in hair, eyes, nose, mouth, colors, \etc while content drives pose, face sizes, where facial parts
are, \etc.  Figure~\ref{fig:comparisons} shows that GNR outperforms other SOTA multimodal I2I frameworks in terms of
both quality and diversity.  GNR produces images with varied colors, hairstyles, eye shapes, facial structures, \etc
while other frameworks differs mostly in colors. Note that the style vectors are chosen at random and the images are not
cherry-picked. We also compared to the very recent work of AniGAN~\cite{li2021anigan} in Fig.~\ref{fig:anigan}. However,
we are only comparing with test cases presented in the paper as the source code is not released. Note that even though
AniGAN is trained on a bigger and more diverse dataset compared to ours, we are able to generate higher quality images
with significantly better diversity. Also, AniGAN generates at $128 \times 128$ resolution while we generate at $256
\times 256$.

\paragraph{Ablation}

Ablation shows the Diversity Discriminator plays a big part in ensuring diverse outputs (Fig.~\ref{fig:ablation}).
Furthermore, replacing the diversity discriminator with a mode seeking loss produces some diversity, but results are
much weaker than GNR. 

\begin{figure}[t!]
    \centering
    \includegraphics[width=1\linewidth]{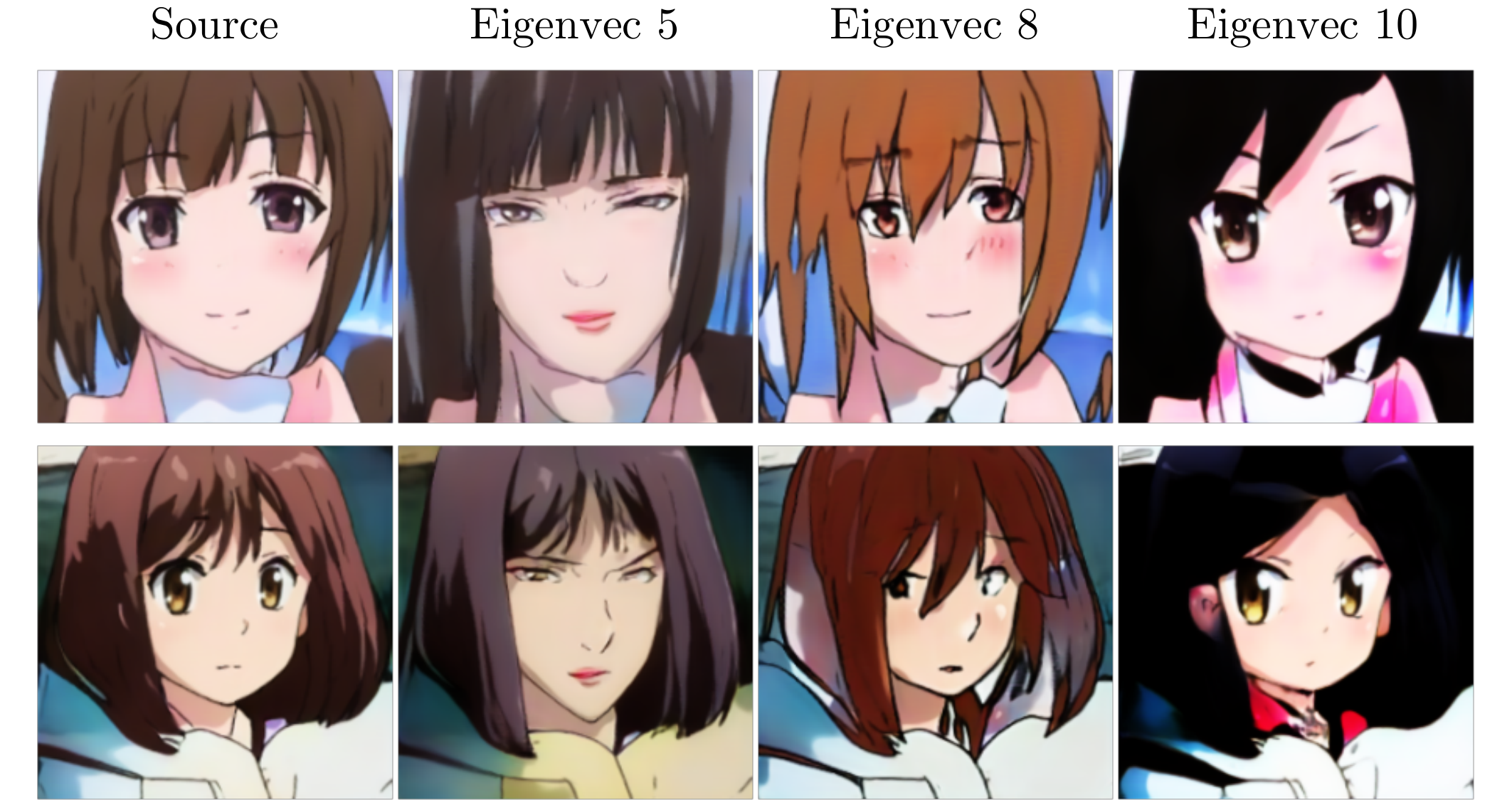}
    \caption{
    {\bf Latent Exploration:} GNR encodes rich semantic information in the style space. We can find meaningful latent direction from SeFa which enables us to perform style editing. Each column has differently sampled style but adds the same latent direction; each eigenvector give us semantically meaningful results. col 1: Small eyes and sharp features; col 2: abstract art lines; col 3: Big eyes and dark hair. }
    \label{fig:sefa}
\end{figure}

\subsection{Quantitative Comparisons}

We evaluate GNR quantitatively in Table~\ref{tab:fid}, using diversity FID, FID , and LPIPS.
Generally, we show that GNR is significantly better than all other SoTA frameworks in all metrics over all experiments
in Table~\ref{tab:fid}. Both DFID and LPIPS focus on diversity of images, and scores in those metrics quantitatively
confirms our diversity is better than other frameworks. For selfie2anime, our FID is only very slightly worse than
CouncilGAN. Note that FID does not represent output diversity as it is a population matching metric computed over all
images. 

\begin{table*}[]
\centering
\begin{tabular}{c|c|c|c|c|c|c}
    &\multicolumn{3}{c}{selfie2anime} &\multicolumn{3}{c}{cat2dog} \\
     \hline
    & DFID $\downarrow$ &FID $\downarrow$ & LPIPS $\uparrow$  & DFID $\downarrow$ &FID $\downarrow$ & LPIPS $\uparrow$ \\ \hline
    GNR        & \textbf{35.6} & \textbf{34.4} &\textbf{0.505} & \textbf{26.1}  &\textbf{26.9} & \textbf{0.569} \\ \hline
    DRIT++     & 94.6 & 63.8 & 0.201 & 160.1 &91.5 & 0.231 \\ \hline
    CouncilGAN & 56.2 & 38.1 & 0.430 & 172.5 & 90.8 & 0.298\\ \hline
    StarGANv2  & 83.0 & 59.8 & 0.427 & 53.6 & 44.2 & 0.530\\ 
\end{tabular}
\caption{\textbf{Quantitative Comparisons:} We compare GNR with other SoTA frameworks. DFID compares the distribution of a single image translated with different styles with the distribution of real images, thus focus on output diversity. FID compares general image quality while LPIPS also focuses on output diversity. Our method shows significant improvements across all metrics, especially on the Diversity FID and LPIPS metric which measures diversity.}
\label{tab:fid}
\end{table*}

\paragraph{FID}
The Fréchet Inception Distance (FID)~\cite{heusel2017gans} is a widely used metric to measure the performance of
generative models. It measures the realism and diversity of generated images compared to real images.  All FID calculations are done using FID$_\infty$~\cite{chong2019effectively}, as the original FID
has a generator dependent bias, and FID$_\infty$ removes this using simple extrapolation procedures.  

\paragraph{Diversity FID}

While previous I2I frameworks~\cite{nizan2020council,lee2018diverse,choi2019stargan2} also use FID for evaluation, they generate one image
per input image -- this does not measure style diversity; for example, a method might only be capable of generating a single unique image per input image and still get a good FID.  Our Diversity FID (DFID) is a simple modification to the  
original FID algorithm that aims to measure style diversity for multi-modal I2I framework.  Recall the assumption and
notation of section~\ref{diversity} -- applying a random augmentation to an image
results in a content code that is a sample from the prior on content code.  This assumption means that if we generate
codes for $M$ augmentations of any given test image, we have $M$ IID samples from $P(C)$.  In turn, if our model of
$P(\hat{Y}|C=c_i)$ is correct, then generating anime from these augmentations using randomly selected style codes should
result in a set of anime that are IID samples from $P(Y)$.   This should be true for {\em any} selected image.  In turn, if the method works,
\begin{equation}
    DFID = \frac{1}{N}\sum_{i=1}^N FID(\{G(T_M(x_i), z) | z \sim \mathcal{N}(0, I) \}, Y)
\end{equation}
should be small.
In all experiments, we use $M=1000$ and $N=100$. All test images used are the same across experiments.

\paragraph{LPIPS}
LPIPS~\cite{zhang2018perceptual} is a metric that computes the perceptual similarity between two images using deep network actiavtions. Following the LPIPS metric used in StarGANv2, we use LPIPS to measure similarity of the different images generated a single source image with varied style vectors. A small LPIPS score indicates mode collapse and a lack of diversity when varying style codes. For each test image, we randomly generate $10$ outputs which we then compute their pairwise LPIPS distance. We then average the distance across all test images, $N=100$. 

\subsection{Video to Video Translation}  
Our definition of style and content means that, when a face moves in a frame, the style should not change but the
content will.  In particular, the content encodes where features are while style encodes what they look like.  In turn,
content codes should capture all frame-to-frame movement, and we should be able to synthesize anime video {\em without
  ever training on temporal sequences}.  We apply GNR to face videos frame by frame, then assemble the resulting frames
into a video. Results in Fig.~\ref{fig:interpolation} row 2 shows that GNR produces images that moves according
to the source while maintaining consistent appearances temporally.   Note that very pleasing visual effects can be obtained by
manipulating what style code is used at what point on the timeline; for example, a synthesized anime could change style
at each beat.   The full video in the Supplementary shows a smooth video to video translation with minimal temporal artifacts.  

We also tested the SOTA in selfie2anime generation CouncilGAN in the same task of V2V as shown in
Fig.~\ref{fig:interpolation}. While CouncilGAN adequately captures the pose of the image, it is not temporally
consistent. Artifacts are present and the shades of the hair changes across time. The interpolation across 2 styles also
showed the lack of image diversity. 

\section{Discussion}
\subsection{Disentanglement}
From our results in Fig.~\ref{fig:results}, we can clearly see that content dictates the pose, face shape, and to some extent, the hairstyle while style controls everything else. The disentanglement between content and style arise from our style consistency loss, which enforces the style codes of randomly augmentations of the same image be consistent. Our Diversity discriminator then forces the distribution of images across styles be diverse while cycle consistency loss ensures information is not loss in the translation.

\subsection{Unsupervised Latent Space Editing}
GANs N' Roses uses the StyleGANv2~\cite{Karras2019stylegan2} architecture which allows latent space editing techniques to also work on GNR. SeFa~\cite{shen2020closedform} finds latent direction that corresponds to large semantic changes in a style-based GAN by finding eigenvectors of the style modulation weight parameters. We show that our style space is highly expressive and editable using these eigenvectors, Fig.~\ref{fig:sefa}. In each row, we have different source image with different intial style code. When we apply the same eigenvector direction to the different style code, the output images exhibit similar style changes. This allows a degree of controllability over the I2I translation.

\section{Conclusion}
In this work, we define content as \emph{where things are} and style as \emph{what they look like} in the setting of multimodal I2I translation. Using this simple defintion, we propose GANs N' Roses, a multimodal I2I framework that produces truly diverse images that captures the complex artistic styles given a single input image. We then show that our defintion of content and style allows GNR to be applied to the difficult problem of video to video translation with no additional training.

{\small
\bibliographystyle{ieee_fullname}
\bibliography{egbib}
}

\end{document}